\documentclass[11pt,a4paper]{article}
\usepackage[hyperref]{acl2021}
\usepackage{times}
\usepackage{latexsym}
\usepackage{graphicx}
\usepackage{lipsum}

\usepackage{float}
\usepackage{algorithm}
\usepackage{algorithmic}
\usepackage{placeins}
\usepackage{subfigure}
\usepackage{booktabs}
\usepackage{amsmath,amssymb}
\usepackage{cleveref}
\usepackage{multirow}
\usepackage{comment}

\usepackage{microtype}
\usepackage{newtxmath}

\aclfinalcopy 
\def\aclpaperid{3829} 

\definecolor{mygreen}{RGB}{25, 148, 7}

\title{A Cluster-based Approach for Improving Isotropy \\in  Contextual Embedding Space}

\author{Sara Rajaee$^1$ \and Mohammad Taher Pilehvar$^{1,2}$ \\
  $^1$ Iran University of Science and Technology, Tehran, Iran \\
  $^2$ Tehran Institute for Advanced Studies, Tehran, Iran \\
  \texttt{sara{\_}rajaee@comp.iust.ac.ir} \\
  \texttt{mp792@cam.ac.uk}}

\date{}

\begin{document}
\maketitle
\begin{abstract}

The representation degeneration problem in Contextual Word Representations (CWRs) hurts the expressiveness of the embedding space by forming an anisotropic cone where even unrelated words have excessively positive correlations.
Existing techniques for tackling this issue require a learning process to re-train models with additional objectives and mostly employ a global assessment to study isotropy. 
Our quantitative analysis over isotropy shows that a local assessment could be more accurate due to the clustered structure of CWRs. 
Based on this observation, we propose a local cluster-based method to address the degeneration issue in contextual embedding spaces.
We show that in clusters including punctuations and stop words, local dominant directions encode structural information, removing which can improve CWRs performance on semantic tasks. 
Moreover, we find that tense information in verb representations dominates sense semantics. 
We show that removing dominant directions of verb representations can transform the space to better suit semantic applications. 
Our experiments demonstrate that the proposed cluster-based method can mitigate the degeneration problem on multiple tasks.\footnote{The code for our experiments is available at \url{https://github.com/Sara-Rajaee/clusterbased_isotropy_enhancement/}}

\end{abstract}

\section{Introduction}
Despite their outstanding performance, CWRs are known to suffer from the so-called \emph{representation degeneration problem} that makes the embedding space anisotropic \citep{DBLP:journals/representation-degproblem}. 
In an anisotropic embedding space, word vectors are distributed in a narrow cone, in which even unrelated words are deemed to have high cosine similarities. This undesirable property hampers the representativeness of the embedding space and limits the diversity of encoded knowledge \cite{ethayarajh-2019-contextual}.

To better understand the representation degeneration problem in pre-trained models, we analyzed the embedding space of GPT-2~\citep{radford2019language}, BERT \citep{devlin-2019-bert}, and RoBERTa ~\citep{DBLP:journals/corr/abs-1907-11692}.
We found that, despite being extremely anisotropic in all non-input layers from a global sight, the embedding space is significantly more isotropic from a local point of view (when embeddings are clustered and each cluster is made zero-mean).
Motivated by this observation and based on previous studies that highlight the clustered structure of CWRs~\citep{visualize-reif, michael-etal-2020-asking}, we extend the technique of \citet{Mu-all-but-the-top} with a further clustering step. 
In our proposal, we cluster embeddings and apply PCA on individual clusters to find the corresponding principal components (PCs) which indicate the dominant directions for each specific cluster.
Nulling out these PCs for each cluster renders a more isotropic space.
We evaluated our cluster-based method on several tasks, including Semantic Textual Similarity (STS) and Word-in-Context (WiC). 
Experimental results indicate that our cluster-based method is effective in enhancing the isotropy of different CWRs, reflected by the significant performance improvements in multiple evaluation benchmarks.

In addition, we provide an analysis on the reasons behind the effectiveness of our cluster-based technique. 
The empirical results show that most clusters contain punctuation tokens, such as periods and commas. 
The PCs of these clusters encode structural information about context, such as sentence style; hence, removing them can improve CWRs performance on semantic tasks. 
A similar structure exists in other clusters containing stop words. 
The other important observation is about verb distribution in the contextual embedding space. 
Our experiments reveal that verb representations are separated across the tense dimension in distinct sub-spaces.
This brings about an unwanted peculiarity in the semantic space: representations for different senses of a verb tend to be closer to each other in the space than the representations for the same sense that are associated with different tenses of the same verb. 
Indeed, removing such PCs improves model's ability in downstream tasks with dominant semantic flavor. \par

\section{Isotropy in CWRs}
Isotropy is a desirable property of word embedding spaces and arguably any other vector representation of data in general \citep{Huang_2018_CVPR,Cogswell-Decorrelation}. 
From the geometric point of view, a space is called isotropic if the vectors within that space are uniformly distributed in all directions. 
Lacking isotropy in the embedding space affects not only the optimization procedure (e.g., model’s accuracy and convergence time) but also the expressiveness of the embedding space; hence, improving the isotropy of the embedding space can lead to performance improvements \citep{Wang2020Improving,10.5555/3045118.3045167}.

We measure the isotropy of embedding space using the partition function of \citet{arora-etal-2016-latent}:
\begin{equation}
    F(u) = \sum_{i = 1}^{N} e^{u^T w_i}\label{eq:1}
\end{equation}
\noindent where $u$ is a unit vector, $w_i$ is the corresponding embedding for the $i^{th}$ word in the embedding matrix $\mathrm{W}\in{\rm I\!R}^\mathrm{N \times D}$, $\mathrm{N}$ is the number of words in the vocabulary, and $\mathrm{D}$ is the embedding size. 
\citet{arora-etal-2016-latent} showed that $F(u)$ can be approximated using a constant for isotropic embedding spaces. Therefore, for the set $U$, which is the set of eigenvectors of $\mathrm{W}^T\mathrm{W}$, in the following equation, $\mathrm{I}(\mathrm{W})$ would be close to one for a perfectly isotropic space \citep{Mu-all-but-the-top}.
\begin{equation}
    \mathrm{I}(\mathrm{W}) = \frac{min_{u\in U}F(u)}{max_{u\in U}F(u)}\label{eq:2}
\end{equation}

\subsection{Analyzing Isotropy in pre-trained CWRs}
Using the above metric, we analyzed the representation degeneration problem globally and locally.

\begin{figure}[t!]
    \centering
    {\includegraphics[width=7.5cm]{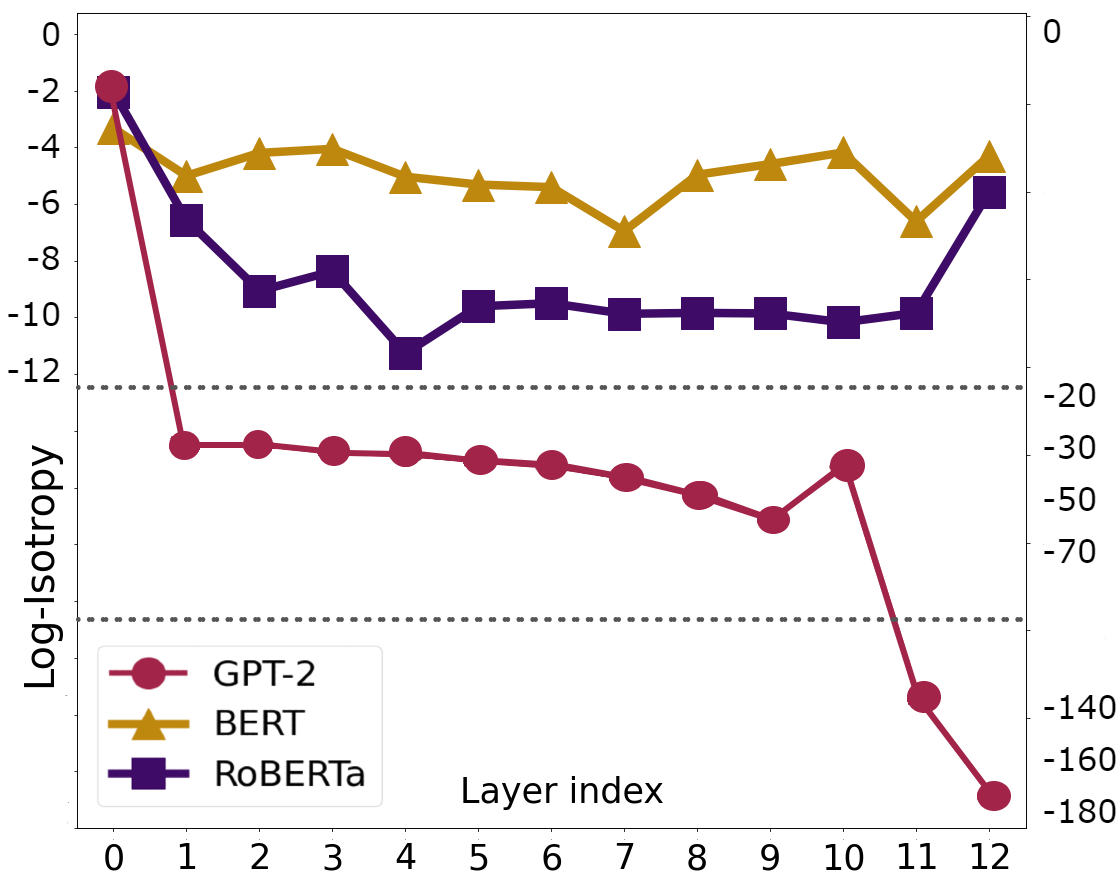}} 
    \caption{Layer-wise isotropy for different CWRs on the STS-B dev set ($\uparrow$ log-isotropy: $\uparrow$ isotropy). Given the large difference, BERT and RoBERTa are shown on the left axis and GPT-2 on the right.}
    \label{fig:fig-iso-alllayers}
\end{figure}

\paragraph{Global assessment.} We quantified isotropy in all layers for GPT-2, BERT, and RoBERTa on the development set of STS-Benchmark \citep{cer-etal-2017-semeval}.
Figure \ref{fig:fig-iso-alllayers} shows the trend of isotropy in all layers based on $\mathrm{I}(\mathrm{W})$. 
Clearly, all CWRs are extremely anisotropic in all non-input layers. 
While the isotropy of GPT-2 decreases consistently in upper layers, that for RoBERTa has a semi-convex form in which the last layer (except for the input layer) has the highest isotropy. 
Also, interestingly, the input layer in GPT-2 is more isotropic than those for the other two models. This observation contradicts with what has been previously reported by ~\citet{ethayarajh-2019-contextual}.

\paragraph{Local assessment.} In the light of the clustered structure of the embedding space in CWRs \citep{visualize-reif}, we carried out a local investigation of isotropy. To this end, we clustered the space using $k$-means and measured isotropy after making each cluster zero-mean \citep{Mu-all-but-the-top}. 
Table \ref{zero-mean-table} shows the results for different number of clusters (each being the average of five runs).
When the embedding space is viewed closely, the distribution of CWRs is notably more isotropic.
Clustering significantly enhances isotropy for BERT and RoBERTa, making their embedding spaces almost isotropic.
However, GPT-2 is still far from being isotropic.
This contradicts with the observation of \citet{cai2021isotropy}. 

\begin{figure}

    \centering
    \includegraphics[width=0.48\textwidth]{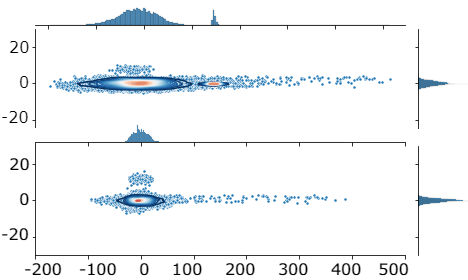} 
    \caption{GPT-2 embeddings on STS-B dev set before (top) and after (bottom) a local zero-mean operation.
    }
    \label{fig:fig-GPT-zeromean}

\end{figure}

A possible explanation for these contradictions is the different metric used by ~\citet{ethayarajh-2019-contextual} and  \citet{cai2021isotropy} for measuring isotropy: cosine similarity.
Randomly sampled words in an anisotropic embedding space should have high cosine similarities (a near-zero similarity denotes isotropy). 
However, there are exceptional cases where this might not hold (an anisotropic embedding space where sampled words have near-zero cosine similarities). 
In Figure \ref{fig:fig-GPT-zeromean}, we illustrate GPT-2 embedding space as an example for such an exceptional cases. 
Making individual clusters zero-mean (bottom) improves isotropy over the baseline (top).
However, the embeddings are still far from being uniformly distributed in all directions.
Instead, they are distributed around a horizontal line.
This leads to a near-zero cosine similarity for randomly sampled words while the embedding space is anisotropic. 
Hence, cosine similarity might not be a proper metric for computing isotropy.

\begin{table}
\centering
\setlength{\tabcolsep}{9pt}
\scalebox{0.85}{
\begin{tabular}{lccc}
\toprule
\textbf{} & \textbf{GPT-2} & \textbf{BERT} & \textbf{RoBERTa} \\ 
\midrule
Baseline & ~~5.02E-174 & 5.05E-05 & 2.70E-06\\ 
\midrule
$k = 1$  & ~~2.49E-220 & 0.010  & 0.015\\
$k = 3$  & 9.42E-66 & 0.040 & 0.290 \\
$k = 6$  & \textbf{1.40E-41} & 0.125 & 0.453\\
$k = 9$  & 1.18E-41 & 0.131 & 0.545\\
$k = 20$  & 4.06E-47 & \textbf{0.262} & \textbf{0.603}\\

\bottomrule
\end{tabular}
}
\caption{\label{zero-mean-table} CWRs isotropy after clustering and making each cluster zero-mean separately (results for different number of clusters ($k$) on STS-B dev set). }
\end{table}

\section{Cluster-based Isotropy Enhancement}

The degeneration problem in the embedding space can be attributed to the training procedure of the underlying models, which are often language models trained through likelihood maximization with the weight tying trick
\citep{DBLP:journals/representation-degproblem}. 
Maximizing the likelihood of a specific word embedding (minimizing that for others) requires pushing it towards the direction of the corresponding hidden state, which results in the accumulation of the learnt word embeddings into a narrow cone. 

Previous work has shown that nulling out dominant directions of an anisotropic embedding space can make the space isotropic and improve its expressiveness \citep{Mu-all-but-the-top}. 
We refer to this as the \textit{global} approach.
This method was proposed for static embeddings. Hence, it might not be  optimal for contextual embeddings, especially in the light that the latter tends to have a clustered structure.
For instance, recent work suggests that word types (e.g., verbs, nouns, punctuations), entities (e.g., personhood, nationalities, and dates), and even word senses \cite{michael-etal-2020-asking,10.1162/coli_a_00405,visualize-reif} create local distinct clustered areas in the contextual embedding space. 
Moreover, our local assessment shows that it is not necessarily the case that all clusters share the same dominant directions. Hence, discarding dominant directions that are computed globally is not efficient for removing local degenerated directions. 
Consequently, it is more logical to have a cluster-specific dropping of dominant directions.

Based on these observations, we propose a cluster-based approach for isotropy enhancement.
Specifically, instead of determining dominant directions globally, we obtain them separately for different sub-spaces and discard for each cluster only the corresponding cluster-specific dominant directions.
To this end, we employ Principal Component Analysis (PCA) to compute local dominant directions in clusters. Geometrically, principal components (PCs) represent those directions in which embeddings have the most variance (maximum elongation).
In our proposed method, we first cluster word embeddings using a simple $k$-means algorithm. 
After making each cluster zero-mean, the top PCs of every cluster are removed separately. Adding a clustering step helps us to eliminate the local dominant directions of each cluster.
We will show in Section \ref{sec-analyze} that different linguistic knowledge is encoded in the dominant directions of various clusters.
Moreover, numerical results show that in comparison with the global approach, our method can make the embedding space more isotropic, even when the fewer number of PCs are nulled out.

\begin{table*}
\centering
\setlength{\tabcolsep}{6pt}
\scalebox{0.78}{
\begin{tabular}{l l c c c c c c c }

\toprule
 & \textbf{\textbf{Model}} & \textbf{STS 2012} & \textbf{STS 2013} & \textbf{STS 2014} & \textbf{STS 2015}  &\textbf{STS 2016} & \textbf{SICK-R} & \textbf{STS-B}  \\ 
\midrule
\multirow{3}{*}{\bf Baseline} &
GPT-2     & 26.49 & 30.25  & 35.74 & 41.25              &  46.40  & 45.05     & 24.8     \\
& BERT-base    & 42.87 & 59.21  & 59.75 & 62.85              &  63.74  & 58.69     & 47.4     \\
& RoBERTa-base & 33.09 & 56.44  & 46.76 & 55.44              &  60.88  & 61.28     & 56.0     \\
\midrule
\multirow{3}{*}{\bf Global approach} &
GPT-2     & 51.42 & 69.71  & 55.91 & 60.35              &  62.12  & 59.22     & 55.7     \\
& BERT-base    & 54.62 & 70.39  & 60.34 & 63.73              &  69.37  & 63.68     & 65.5     \\
& RoBERTa-base & 51.59 & 73.57  & 60.70 & 66.72              &  \textbf{69.34}  & 65.82     & 70.1     \\
\midrule
\multirow{3}{*}{\bf Cluster-based approach} &
GPT-2       & \textbf {52.40} & \textbf{72.71}  & \textbf{59.23} & \textbf{62.19}                    & \textbf {64.26} & \textbf{59.51}  & \textbf{62.3}     
\\
& BERT-base      & \textbf{58.34}  & \textbf{75.65}  & \textbf{63.55} & \textbf{64.37}                    & \textbf{69.63}  & \textbf{63.75}  & \textbf{66.0}     
\\
& RoBERTa-base   & \textbf{54.87}  & \textbf{76.70}  & \textbf{64.18} & \textbf{67.05}                    & {69.28}  & \textbf{66.93}  & \textbf{71.4}      
\\
\bottomrule

\end{tabular}
}
\caption{\label{experiment-tabel} Spearman correlation performance of three pre-trained models (baseline) on the Semantic Textual Similarity datasets, before and after isotropy enhancement with the global and cluster-based (our) approach.}
\end{table*}

\begin{table*}
\centering
\setlength{\tabcolsep}{16.5pt}
\scalebox{0.78}{
\begin{tabular}{l c c c c c c c}
\toprule
  & \textbf{RTE} & \textbf{CoLA} & \textbf{SST-2} & \textbf{MRPC}  &\textbf{WiC} & \textbf{BoolQ} & \textbf{\it Average}\\ 
\midrule
\multirow{1}{*}{\bf Baseline} &
            54.4 & 38.0 & 80.1 & 70.2 &  60.0  & 64.7  &  61.2   \\

\multirow{1}{*}{\bf Global approach} &
            56.2  & 38.8  & 80.2 & 72.1 &  60.7  &  64.9 & 62.1  \\

\multirow{1}{*}{\bf Cluster-based approach} &
            \textbf{56.5}  & \textbf{40.7}  & \textbf{82.5} & \textbf{72.4} &  \textbf{61.0}  &  \textbf{66.4} &  \textbf{63.2}  \\
\bottomrule

\end{tabular}
}
\caption{\label{experiment-tabel-classification} Results on the classification tasks (BERT) in terms of accuracy (except for CoLA: Matthew's correlation). }
\end{table*}

\section{Experiments}
We carried out experiments on the following benchmarks.
As for Semantic Textual Similarity (STS), which is the main benchmark for our experiments, we experimented with STS 2012-2016 datasets \citep{agirre-etal-2012-semeval, agirre-etal-2013-sem, agirre-etal-2014-semeval, agirre-etal-2015-semeval, agirre-etal-2016-semeval}, the SICK-Relatedness dataset (SICK-R) \citep{marelli-etal-2014-sick}, and the STS benchmark (STS-B).
For the STS task, we report results for GPT-2, BERT, and RoBERTa.
We also experimented with a number of classification tasks: Recognizing Textual Entailment from the GLUE benchmark \citep[RTE]{wang-etal-2018-glue}, the Corpus of Linguistic Acceptability \citep[CoLA]{warstadt-etal-2019-neural}, Stanford Sentiment Treebank \citep[SST-2]{socher-etal-2013-recursive}, Microsoft Research Paraphrase Corpus \citep[MRPC]{dolan-brockett-2005-automatically}, Word-in-Context \cite[WiC]{pilehvar-camacho-collados-2019-wic}, and BoolQ \cite{clark-etal-2019-boolq}.
For the classification tasks, we limit our experiments to BERT and extract features to train an MLP.
Further details on the datasets and system configuration can be found in Appendix \ref{app-configurations}.

We benchmark our cluster-based approach with the pre-trained CWRs (baseline) and the global method.
As it was mentioned before, this method is similar to ours in its elimination of a few top dominant directions but with the difference that these directions are computed globally (in contrast to our local cluster-based computation).
The best setting for each model is selected based on performance on the STS-B dev set. 
The reported results are the average of five runs.

\subsection{Results}
Tables \ref{experiment-tabel} and \ref{experiment-tabel-classification} report experimental results.
As can be seen, globally increasing isotropy can make a significant improvement for all the three pre-trained models. 
However, our cluster-based approach can achieve notably higher performance compared to the global approach.
We attribute this improvement to our cluster-specific discarding of dominant directions. 
Both global and cluster-based methods null out the optimal number of top dominant directions (tuned separately, cf. Appendix \ref{app-configurations}), but the latter identifies them based on the specific structure of a sub-region in the embedding space (which might not be similar to other sub-regions).

\begin{table*}[!t]
\centering
\scalebox{0.8}{
\setlength{\tabcolsep}{12pt}
\begin{tabular}{l  cccc cccc}
\toprule
\multicolumn{1}{c}{} &
\multicolumn{4}{c}{\textbf{Baseline}} & 
\multicolumn{4}{c}{\textbf{Removed PCs}}\\

\cmidrule(lr){2-5}
\cmidrule(lr){6-9}
\textbf{\textbf{Model}} & \textbf{ST-SM} & \textbf{ST-DM} & \textbf{DT-SM} & \textit{Isotropy}  &\textbf{ST-SM} & \textbf{ST-DM} & \textbf{DT-SM}& \textit{Isotropy}   \\ 
\midrule
GPT-2   & 48.82 & 48.19  & 50.86 & 2.26E-05 &  ~~9.32  & ~~9.53  & ~~9.49  & 0.17\\
BERT    & 13.44 & 14.24  & 14.87 & 2.24E-05 &  10.31 & 10.50 & 10.32 & 0.32 \\
RoBERTa & ~~5.89  & ~~6.31   & ~~6.86  & 1.22E-06 &  ~~4.78  & ~~5.00  & ~~4.89  & 0.73 \\

\bottomrule
\end{tabular}
}
\caption{\label{tense-bias-table}The mean Euclidean distance of a sample occurrence of a verb to all other occurrences of the same verb with the Same-Tense and the Same-Meaning (ST-SM), the Same-Tense but Different-Meaning (ST-DM), and a Different-Tense but the Same-Meaning (DT-SM). Semantically, it is desirable for DT-SM to be lower than ST-DM.}
\end{table*}

\section{Discussion}
\label{sec-analyze}
In this section, we provide a brief explanation for reasons behind the effectiveness of the cluster-based approach through investigating the 
linguistic knowledge encoded in the dominant local directions. We also show that enhancing isotropy reduces convergence time.

\subsection{Linguistic knowledge}

\begin{figure}[t!]
\centering
    {\includegraphics[width=0.45\textwidth]{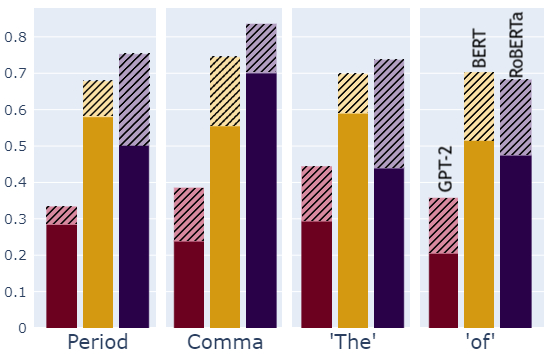}} 
    \caption{The percentage of nearest neighbours that share similar structural and syntactic knowledge, before (lighter, pattern-filled) and after removing dominant directions in pre-trained CWRs. }
    \label{fig:fig-punc}
\end{figure}

\textbf{Punctuations and stop words.}
We observed that local dominant directions for the clusters of punctuations and stop words carry structural and syntactic information about the sentences in which they appear. For example, the two sentences ``A man is crying.'' and ``A woman is dancing.'' from STS-B do not have much in common in terms of semantics but are highly similar with respect to their style.
To quantitatively analyze the distribution of this type of tokens in CWRs, we designed an experiment based on the dataset created by \citet{ravfogel-etal-2020-unsupervised}. 
The dataset consists of groups in which sentences are structurally and syntactically similar but have no semantic similarity. 
We picked 200 different structural groups in which each group has six semantically different sentences. Then, using the $k$-NN algorithm, we calculated the percentage of nearest neighbours which are in the same group before and after removing local dominant directions. 
We evaluated this for period and comma, which are the most frequent punctuations, and “the” and “of” as the most contextualized stop words \citep{ethayarajh-2019-contextual}. 
The reported results in Figure \ref{fig:fig-punc} show that the representations for punctuations and stop words are biased toward structural and syntactic information of sentences; hence, removing their dominant directions reduces the number of same-group nearest neighbours.
The improvement from our local isotropy enhancement can be partially attributed to attenuating this type of bias.

\paragraph{Verb Tense.}
Our experiments show that tense is more dominant in verb representations than sense-level semantic information. 
To have a precise examination of this hypothesis, we used SemCor \citep{miller-etal-1993-semantic}, a dataset comprising around 37K sense-annotated sentences. We collected representations for polysemous verbs that had at least two senses occurring a minimum of 10 times.
Then, for each individual verb, we calculated Euclidean distance to the contextual representation of the same verb: (1) with the same tense and the same meaning, (2) with the same tense but a different meaning, and (3) with a different tense and the same meaning. 
The experimental results reported in Table \ref{tense-bias-table} confirm the hypothesis and show the effectiveness of the cluster-based approach in bringing together verb representations that correspond to the same sense, even if they have different tense.

\begin{figure}[t!]
    \centering
    {\includegraphics[width=0.45\textwidth]{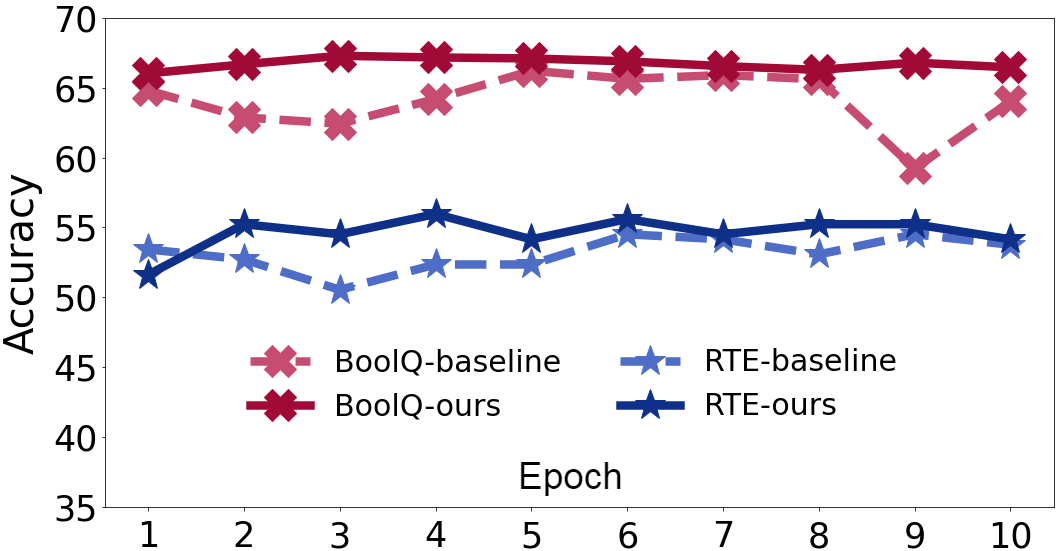}} 
    \caption{The impact of our cluster-based isotropy enhancement on per-epoch performance for two tasks.}
    \label{fig:fig-convergance}
\end{figure}

\subsection{Convergence time}
In the previous experiments, we showed that the contextual embeddings are extremely anisotropic and highly correlated. Such embeddings can slow down the learning process of deep neural networks.
Figure \ref{fig:fig-convergance} shows the trend of convergence for the BoolQ and RTE tasks (dev sets). 
By decreasing the correlation between embeddings, our method can reduce convergence time.

\section{Conclusions}

In this paper, we proposed a cluster-based method to address the representation degeneration problem in CWRs. 
We empirically analyzed the effect of clustering and showed that, from a local sight, most clusters are biased toward structural information. 
Moreover, we found that verb representations are distributed based on their tense in distinct sub-spaces. 
We evaluated our method on different semantic tasks, demonstrating its effectiveness in removing local dominant directions and improving performance. 
As future work, we plan to study the effect of fine-tuning on isotropy and on the encoded linguistic knowledge in local regions.

\bibliographystyle{acl_natbib}
\bibliography{anthology,acl2021}

\newpage
\newpage
\newpage

\appendix

\section{Isotropy statistics}
Table ~\ref{isotropy-all-layers-table} shows isotropy statistics for GPT-2, BERT, and RoBERTa. GPT-2's embedding space is extremely anisotropic in upper layers. 
Hence, more PCs are required to be eliminated to make this embedding space isotropic in comparison to BERT and RoBERTa, both in the cluster-based approach and the global one \cite{Mu-all-but-the-top}. Also, in almost all layers, BERT has higher a isotropy than RoBERTa.
\FloatBarrier
\begin{table}[ht!]
\setlength{\tabcolsep}{12pt}
\scalebox{0.85}{
\begin{tabular}{l | ccc}
\toprule
 \textbf{\textbf{Model}} & \textbf{GPT-2} & \textbf{BERT} & \textbf{RoBERTa} \\ 
\midrule
layer 0   & 1.5E-02          & 4.6E-04  & 9.1E-03  \\
\midrule
layer 1   & 9.9E-24          & 9.9E-06  & 2.7E-07  \\
layer 2   & \textbf{2.8E-23} & 6.3E-05  & 8.7E-10  \\
layer 3   & 6.1E-26          & 8.8E-05  & 4.2E-09  \\
layer 4   & 1.6E-27          & 9.2E-06  & 5.4E-12  \\
layer 5   & 3.0E-30          & 4.8E-06  & 2.4E-10  \\
layer 6   & 1.6E-32          & 3.9E-06  & 3.1Ef-10  \\
layer 7   & 1.3E-37          & 1.1E-07  & 1.3E-10  \\
layer 8   & 3.4E-45          & 1.0E-05  & 1.4E-10  \\
layer 9   & 6.4E-55          & 2.5E-05  & 1.3E-10  \\
layer 10  & 4.1E-32          & 6.9E-05  & 6.7E-11  \\
layer 11  & ~~1.8E-132         & 2.4E-07  & 1.4E-10  \\
layer 12  & ~~5.0E-174         & \textbf{5.0E-05}  & \textbf{2.7E-06}  \\

\bottomrule
\end{tabular}
}
\caption{\label{isotropy-all-layers-table} Per-layer isotropy on the STS-B dev set. Numbers have been calculated based on $\mathrm{I}(\mathrm{W})$. }
\end{table}
\FloatBarrier

\section{Experimental Setup}
\label{app-configurations}
\subsection{Dataset details}
\paragraph{STS.} 
In the Semantic Textual Similarity task, the provided labels are between 0 and 5 for each paired sentence. We first calculate sentence embeddings by averaging all word representations in each sentence and then compute the cosine similarity between two sentence representations as a score of semantic relatedness of the pair. 

\paragraph{RTE.}
The Recognizing Textual Entailment dataset is a classification task from the GLUE benchmark \citep{wang-etal-2018-glue}. Paired sentences are collected from different textual entailment challenges and labeled as \emph{entailment} and \emph{not-entailment}.

\paragraph{CoLA.} 
The Corpus of Linguistic Acceptability \citep{warstadt-etal-2019-neural} is a binary classification task in which sentences are labeled whether they are grammatically acceptable. 

\paragraph{SST-2.} 
The Stanford Sentiment Treebank \citep{socher-etal-2013-recursive} is a binary sentiment classification task. 

\paragraph{MRPC.}
The Microsoft Research Paraphrase Corpus \citep{dolan-brockett-2005-automatically} consists of paired sentences, and the goal is determining whether, in a pair, sentences share similar semantics or not.

\paragraph{WiC.}
Word-in-Context \citep{pilehvar-camacho-collados-2019-wic} is a binary classification task in which it should be determined if a target word in two different contexts refers to the same meaning.

\paragraph{BoolQ.}
Boolean Questions \citep{clark-etal-2019-boolq} is a Question Answering classification task. Every sample includes a passage and a yes/no question about the passage. 

\subsection{Configurations}
For the classification tasks, we trained a simple MLP on the features extracted from BERT.
The proposed cluster-based approach has two hyperparameters: the number of clusters and the number of PCs to be removed. We selected both of them from range [5, 30] and tuned them on the STS-B dev set. 
In the cluster-based approach,The optimal number of clusters for GPT-2, BERT, and RoBERTa are respectively 10, 27, and 27.
For BERT and RoBERTa, 12 top dominant directions have been removed, while the number is 30 for GPT-2 regarding its extremely anisotropic embedding space.
The tuning of the number of PCs to be eliminated in the global method has been done similarly to the cluster-based approach (on the STS-B dev set): 30, 15, and 25 for GPT-2, BERT, and RoBERTa, respectively.

\section{Isotropy on STS datasets}
In Table \ref{experiment-isotropy-tabel}, we present the isotropy of the contextual embedding spaces calculated using $\mathrm{I}(\mathrm{W})$ on the STS benchmark. The results reveal the effectiveness of the proposed method in enhancing the isotropy of the embedding space.

\begin{table*}[ht!]
\centering
\setlength{\tabcolsep}{9pt}
\scalebox{0.78}{
\begin{tabular}{l c c c c c c c}
\toprule
\textbf{\textbf{Model}} & \textbf{STS 2012} & \textbf{STS 2013} & \textbf{STS 2014} & \textbf{STS 2015}  &\textbf{STS 2016} & \textbf{SICK-R} & \textbf{STS-B}  \\ 
\midrule
        &       &        &       & \emph{Baseline} &         &       &       \\
        \midrule
GPT-2   & ~~1.4E-178 & ~~1.0E-170  & ~~1.4E-172 & ~~2.9E-177            &  ~~6.0E-174  & ~~9.9E-140     & ~~2.6E-105     \\
BERT    & 3.1E-05 & 1.9E-04    & 2.6E-04 & 3.7E-07              &  2.8E-04  & 4.2E-05     & 1.1E-04     \\
RoBERTa & 3.1E-06 & 3.1E-07    & 3.8E-06 & 3.8E-06              &  3.5E-06  & 3.7E-07     & 2.9E-06     \\
\midrule
\midrule
        &       &        &       & \emph{Global approach}&      &       &       \\
        \midrule
GPT-2   & 0.57 & 0.40    & 0.05 & 0.12              &  0.60  & 0.57     & 0.51     \\
BERT    & 0.48    & 0.41    & 0.55 & 0.72              &  0.65  & 0.63     & 0.58     \\
RoBERTa & 0.67 & 0.87    & 0.87 & 0.84              &  0.85  & 0.90     & 0.88     \\
\midrule
\midrule
        &       &        &       & \emph{Cluster-based approach} &       &        &        \\
        \midrule
GPT-2   & \textbf{0.71} & \textbf{0.74}  & \textbf{0.47} & \textbf{0.74}              &  \textbf{0.74} & \textbf{0.78}      & \textbf{0.70}      \\
BERT    & \textbf{0.68} & \textbf{0.61}  & \textbf{0.77} & \textbf{0.81}              &  \textbf{0.75} & \textbf{0.82}      & \textbf{0.73}      \\
RoBERTa & \textbf{0.89} & \textbf{0.91}  & \textbf{0.93} & \textbf{0.92}              &  \textbf{0.89} & \textbf{0.94}      & \textbf{0.90}      \\
\bottomrule
\end{tabular}
}
\caption{\label{experiment-isotropy-tabel}Isotropy of CWRs on multiple STS datasets calculated based on $\mathrm{I}(\mathrm{W})$; a higher value indicates a more isotropic embedding space. Our cluster-based method significantly increases the isotropy of embedding space on all datasets.}
\end{table*}
\par

\section{Word frequency bias in CWRs}
\begin{figure*}
    \centering
    \subfigure[GPT-2 - Baseline]{\includegraphics[width=5cm,height=4cm]{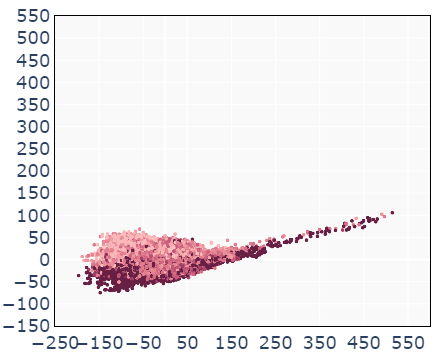} \label{fig:GPT-baseline}} 
    \subfigure[GPT-2 - Global approach]{\includegraphics[width=5cm,height=4cm]{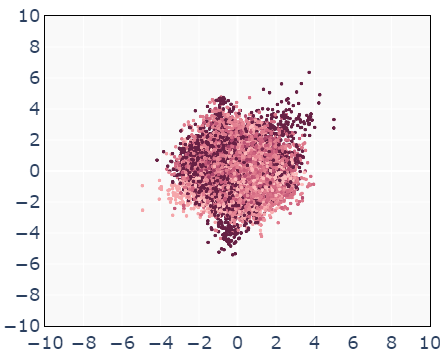}\label{fig:GPT-Noclustering}} 
    \subfigure[GPT-2 - Cluster-based approach]{\includegraphics[width=5cm,height=4cm]{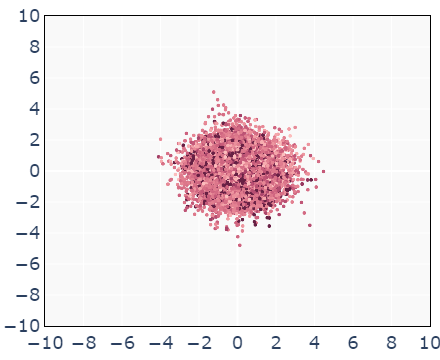}\label{fig:GPT-clustering}}
    
   \subfigure[BERT - Baseline]{\includegraphics[width=5cm,height=4cm]{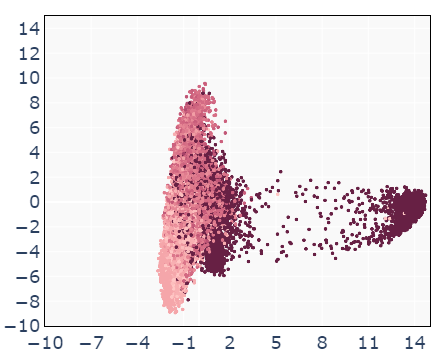}\label{fig:BERT-baseline}} 
    \subfigure[BERT - Global approach]{\includegraphics[width=5cm,height=4cm]{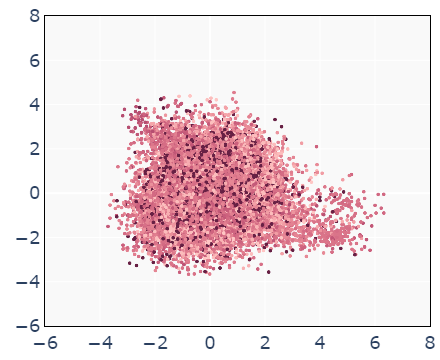} \label{fig:BETR-Noclustering}} 
    \subfigure[BERT - Cluster-based approach]{\includegraphics[width=5cm,height=4cm]{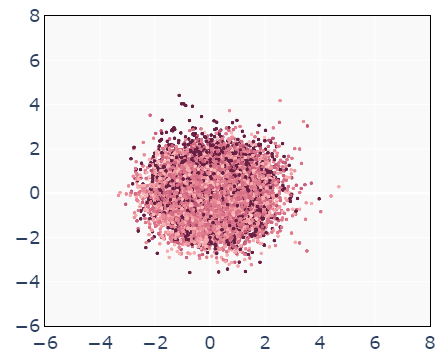}\label{fig:BERT-clustering}}

    \subfigure[RoBERTa - Baseline]{\includegraphics[width=5cm,height=4cm]{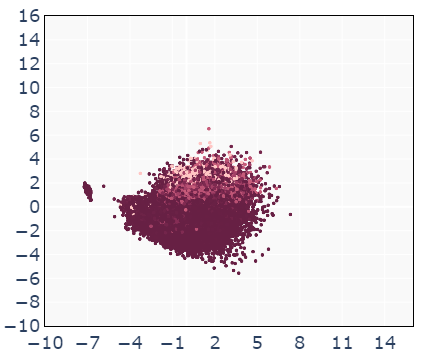}\label{fig:RoBERTa-baseline}} 
    \subfigure[RoBERTa - Global approach]{\includegraphics[width=5cm,height=4cm]{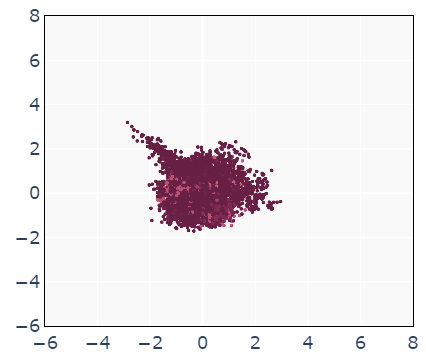} \label{fig:RoBERTa-Noclustering}} 
    \subfigure[RoBERTa - Cluster-based approach]{\includegraphics[width=5 cm,height=4cm]{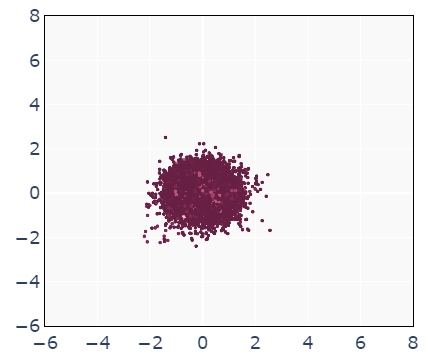} \label{fig:RoBERTa-clustering}}

    \caption{Contextual Word Representations visualization using PCA on STS-B dev set. Colors indicate word frequency in the Wikipedia dump (the lighter point, the more frequent).}
    \label{fig:fig-frequency}
    
\end{figure*}

CWRs are biased towards their frequency information, and words with similar frequency create local regions in the embedding space \citep{Gong-frequency,li-etal-2020-sentence}. 
From the semantic point of view, this is certainly undesirable given that words with similar meanings but different frequencies could be located far from each other in the embedding space. 
This phenomenon can be seen in Figure ~\ref{fig:fig-frequency}.
The encoded knowledge in the local dominant directions partly correspond to frequency information. The embedding space visualization reveals that our approach performs a decent job in removing frequency bias in pre-trained models.

\end{document}